# Integrating GAN and Texture Synthesis for Enhanced Road Damage Detection

First A. TENGYANG CHEN, Second B. JIANGTAO REN

*Abstract*— In the domain of traffic safety and road maintenance, precise detection of road damage is crucial for ensuring safe driving and prolonging road durability. However, current methods often fall short due to limited data. Prior attempts have used Generative Adversarial Networks to generate damage with diverse shapes and manually integrate it into appropriate positions. However, the problem has not been well explored and is faced with two challenges. First, they only enrich the location and shape of damage while neglect the diversity of severity levels, and the realism still needs further improvement. Second, they require a significant amount of manual effort. To address these challenges, we propose an innovative approach. In addition to using GAN to generate damage with various shapes, we further employ texture synthesis techniques to extract road textures. These two elements are then mixed with different weights, allowing us to control the severity of the synthesized damage, which are then embedded back into the original images via Poisson blending. Our method ensures both richness of damage severity and a better alignment with the background. To save labor costs, we leverage structural similarity for automated sample selection during embedding. Each augmented data of an original image contains versions with varying severity levels. We implement a straightforward screening strategy to mitigate distribution drift. Experiments are conducted on a public road damage dataset. The proposed method not only eliminates the need for manual labor but also achieves remarkable enhancements, improving the mAP by 4.1% and the F1-score by 4.5%.

## I. INTRODUCTION

With the continuous advancement of urbanization, road traffic safety and maintainability have become increasingly critical. Within road maintenance, timely detection and accurate identification of road damage are paramount to ensuring smooth traffic flow and driving safety. However, traditional road damage detection methods are often constrained by factors such as limited data and the intricate variability of road environments, making it challenging to meet the requirements of efficient and precise detection.

In recent years, deep learning technology has made remarkable strides in the field of computer vision, many deep learning-based road damage detection methods have been proposed. Zhang *et al.* propose the CrackNet [1] to automating the detection of pavement cracks with pixel-level precision on 3D asphalt surfaces. Leveraging the capabilities of a deep-learning network, the study aims to accurately identify and localize cracks by analyzing intricate patterns and features in large-scale 3D data. For normal image format data, the road damage detection task can be regarded as a typical object detection task, which is one of the fundamental tasks in the field of computer vision. Notably, object detection techniques have gained significant attention due to their extensive practical applications. Nowadays, many works have attempted to apply object detection models, such as YOLOv5 [2], faster-RCNN [3], SSD [4], etc. for road damage detection and achieved promising performance, indicating that object detection models can be used to detect road damage quickly and effectively, and further demonstrating the advantages of deep learning technology in automatic road damage detection. YOLOv5 [5], as an emerging representative in the domain of object detection, has garnered attention for its efficient real-time capabilities and superior detection accuracy. Nonetheless, despite the impressive performance of YOLOv5 in object detection, the task of road damage detection still grapples with the challenge of inadequate data.

To overcome the data scarcity challenge, in recent years, there has been a surge in the development of few-shot learning methods based on meta-learning to address the issue of data scarcity. For instance, in [30], a study employed a rare-disease weighted branch in conjunction with specific data augmentation techniques, effectively resolving cross-domain challenges. On the other hand, [31] utilized Ghost Attention and proposed a feature metric module to eliminate redundant information and measure proposed features. These methods typically undergo pre-training on a diverse set of base classes, followed by fine-tuning on a small number of novel classes to enhance model performance under extremely limited training data conditions. These meta-learning-based methods have demonstrated excellent performance when dealing with scenarios of extremely limited data (less than 30 samples per class). When the data volume falls within the range of several hundred to one or two thousand, employing data augmentation methods to augment the sparsely represented classes is also a viable option.

Regarding data augmentation methods, some studies such as AutoAugment (AA) [36] and Deep AutoAugment (Deep AA) [37] have proposed automated approaches to explore the data augmentation space in order to select the most suitable

This work was supported in part by the Key R&D projects of Guangdong Province under Grant 2022B0101070002. *Corresponding author: Jiangtao Ren.*

First A. TENGYANG CHEN is with the School of Computer Science and Engineering, Sun Yat-Sen University, Guangzhou, 510275 China (e-mail: chenty26@mail2.sysu.edu.cn).

Second B, JIANGTAO REN is with the School of Computer Science and Engineering, Sun Yat-Sen University, Guangzhou, 510275 China (e-mail: issrjt@mail.sysu.edu.cn).

combination of augmentation strategies for a given training dataset. However, these methods often incur high training costs. In contrast, the currently most cost-effective, effective, and practical data augmentation method is TrivialAugment (TA) [32]. TA does not require intricate exploration and tuning but instead uniformly samples a data augmentation function and an intensity value, then applies the augmentation to the image directly. Despite its simplicity, TA has achieved state-of-the-art results on many datasets. Similar to TA, UniformAugment (UA) [38] is also a parameter-free, low-cost data augmentation method. Indeed, the mentioned data augmentation methods often involve combinations of simple image processing functions such as translation, cropping, flipping, and rotation. While effective for general image classification tasks, tasks that require attention to fine-grained features and structures, such as road defect detection, may necessitate more specialized and refined data augmentation methods.

Generative models such as Generative Adversarial Network (GAN) [6] have showcased substantial potential in augmenting image data for task with higher granularity. Through adversarial training, GAN can generate realistic samples, providing richer and more diverse data for model training. In the field of computer vision, numerous endeavors have harnessed GAN for tasks such as image classification and object detection including road damage detection, which have yielded substantial enhancements. For example, Maeda *et al* [7] employed Progressive Growing of GAN (PGGAN) [8] to generate damage with different shapes and seamlessly integrate it onto the appropriate positions of damage-free road surfaces through manual assessment and filtering. This process was complemented by expert validation and Turing tests to curate a collection of high-quality synthetic data. Zhang *et al* [8] employed Conditional Generative Adversarial Network (CGAN) [9] to generate diverse categories of damage, which were subsequently embedded back into the origin images based on annotations. Through manual filtration, inadequately synthesized data were removed, and the refined synthetic dataset was integrated with the original dataset to facilitate expansion. However, the problem has not been well explored and is faced with two critical challenges. First, to achieve a natural embedding, they must select the generated damage that best matches the background, which often has a similar severity to the original damage and the realism still needs further improvement because it cannot be guaranteed that they will be lucky enough to pick the perfect damage that seamlessly blends with the background. As a result, they only enrich the horizonal diversity of damage such as the location and shape of the damage. However, in addition to horizontal diversity, different road damage may also vary in severity, which is referred to as vertical diversity in this paper, as shown in Fig. 1. Some damage may only cause minor surface breakage, which is difficult to be detected by the model, making it a challenging task for detection models. While others can be more severe with obvious damage features. We argue that expanding the vertical diversity of damage is also crucial for improving the detection performance of the model. Second, their process of augmenting data still requires a significant degree of human involvement to ensure the quality of the synthesized data. In our opinion, an ideal data augmentation should serve as a fully automated tool.

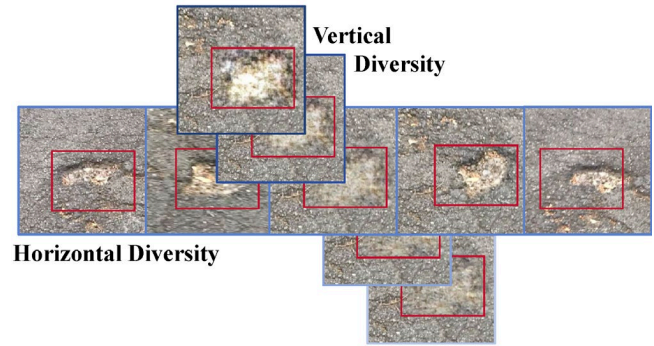

**Figure. 1** The horizontal diversity and vertical diversity of the damage. From the bottom to the top, the severity of the damage gradually increases. This paper aims to enhance the authenticity of synthetic data while extending its vertical diversity.

To address these two challenges, we propose an innovative approach to maximize the quality of synthetic data and extend the vertical diversity while minimizing manual intervention. For the first challenge, in addition to using a simplified WGAN-GP [10] for generating realistic damage ROI with diverse shapes, we further employ a texture synthesis model [11] to effectively extract textures from the road. Mixing the generated ROI with the extracted textures at varying weights enables the synthesized damage with different severity levels and a better fit with the background. Simultaneously, the Poisson blending method [12] ensures a seamless integration of synthesized samples with origin images, making the edge transition more natural and thus preserving the realism of augmented samples. For the second challenge, to save labor costs, we leverage structural similarity for automated sample selection during embedding, liberating the entire process from human involvement. After processing, each original training data has three different augmentation versions with mild, moderate and severe. To prevent distribution shift, for each training data, we randomly select one version to include in the training set. **Fig. 2** illustrates a comparison between our synthesized images and those synthesized by others. It is evident that our synthesized images exhibit higher realism and can depict varying levels of severity. To our knowledge, this is the first method that leverages texture synthesis to enhance the quality and vertical diversity of generated samples.

The contribution of this study is as follows:
(1) We innovatively combine texture synthesis with adversarial generation, resulting in a more natural synthesis of road damage data. Furthermore, it allows control over the severity of damage in the synthesized data, contributing to enhancing the vertical diversity of the dataset.

(2) Compared to other GAN-based approaches, our proposed data augmentation method significantly reduces manual labor costs while remaining easily adaptable and extendable to object detection tasks sharing similar characteristics with road damage detection. In comparison to the original dataset, our method demonstrated a 4.1% improvement in mAP and a 4.5% improvement in F1-score. Compared to existing methods that require manual intervention, our improvement rate is 2.1% higher than their best results. This introduces a novel avenue in the field of data augmentation.

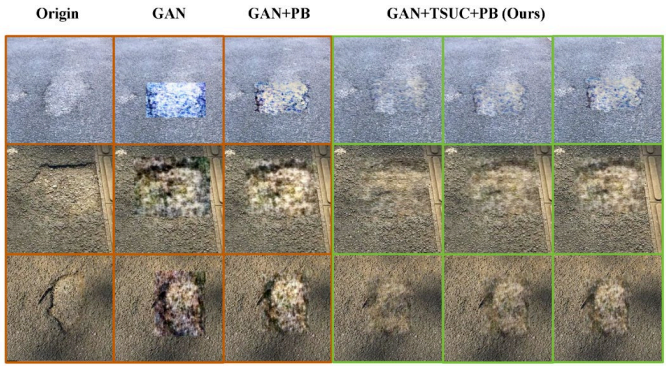

**Figure. 2** Comparison of new data synthesized by different data augmentation methods. The first column is the original images. The second column is directly pasting the machine-selected generated ROIs back into the original images; the third column is embedding the machine-selected generated ROIs back into the original images using PB (Poisson blending); Columns from four to six are the result of mixing the machine-selected generated ROIs with the road textures generated by the TSUC (Texture Synthesis Using CNN) with different weight coefficients, and then embedding using PB.

(3) We employed a straightforward approach to incorporate the synthesized data with varying severity levels into the training set through random selection, effectively mitigating distribution shift issues.

## II. RELATED WORK

### A. Road Damage Detection

Significant progress has been made in the field of road damage detection over the past few years, with many studies focusing on utilizing computer vision and deep learning techniques to enhance the accuracy and efficiency of damage detection. Zhang *et al.* propose an efficient convolutional neural network architecture named CrackNet [1] for automatic crack detection on 3D asphalt road surfaces, with the goal of achieving pixel-level high accuracy. Unlike traditional CNNs, CrackNet omits pooling layers and employs a novel technique to maintain image width and height consistently across all levels, ensuring precision. Comprising five layers and over a million trainable parameters, CrackNet employs proposed linear filters to generate feature maps from input data extracted by feature extractors. Its output is a collection of predicted class scores for all pixels. During testing, CrackNet achieved high accuracy (90.13%), recall (87.63%), and F-measure (88.86%). Compared to traditional machine learning and image algorithms, CrackNet excels particularly in terms of F-measure. Furthermore, leveraging parallel computing techniques, CrackNet seamlessly integrates with data collection software, showcasing effective practicality. This study introduces an efficient and precise solution for 3D road crack detection.

3D images encompass a wealth of structural depth information, rendering them particularly advantageous for road damage detection. However, they impose higher hardware demands on the acquisition equipment. Consequently, an alternate mainstream approach in road damage detection involves directly detect damage on road surface images captured by conventional cameras. This translates it into an object detection task. Models such as YOLOv5 [5], Faster-RCNN [13], and SSD [7] have all been experimented with for road damage detection, yielding promising accuracy. This further validates the feasibility of conducting road damage detection directly using conventional images. Maeda *et al.* [14] first introduced a large-scale Road Damage Dataset (RDD), comprising 9,053 road damage images captured using smartphones mounted on vehicles. These images encompass 15,435 manually annotated instances of road damage. However, the original dataset still suffered from issues like incomplete annotations and imbalanced class distribution. Presently, this dataset has been revamped into RDD2022 [15], addressing these concerns through the collection of additional data and substantial manual efforts for re-annotation. RDD2022 now stands as a comprehensive large-scale dataset, encompassing 47,420 road images collected from six countries including Japan, India, the Czech Republic, Norway, the United States, and China. It comprises over 55,000 meticulously annotated instances of road damage, offering a more refined and improved resource. Empirical evidence demonstrates that as this dataset becomes increasingly comprehensive, the training outcomes of models become more satisfactory [5][13]. This direct correlation underscores the significant impact of data quality and abundance on model training outcomes. However, it is worth noting that the collection and re-annotation process incur substantial human resource costs.

### B. Generative adversarial network

GAN [6] are a deep learning architecture introduced by Ian Goodfellow and his colleagues in 2014. The core idea of GAN involves two neural network models, the Generator and the Discriminator, engaged in a game-like competition to achieve a dynamic equilibrium between generation and discrimination. The Generator's role is to produce realistic data, such as images or audio, from random noise. It essentially functions as a mapping function, transforming input noise into outputs resembling real data. The Generator gradually learns the distribution of real data, enabling it to generate highly convincing synthetic data. The Discriminator's task is to assess whether input data is real or generated by the Generator. It operates as a binary classifier, attempting to correctly classify input as real or generated data. The Discriminator aims to enhance its accuracy in distinguishing real from fake data. During training, the Generator and Discriminator compete against each other. The Generator tries to produce authentic-looking data to deceive the Discriminator, while the Discriminator strives to accurately differentiate between real and generated data. In the training process, the Generator improves its performance by minimizing the probability that generated data is classified as fake. On the other hand, the Discriminator enhances its performance by maximizing the probability of correctly classifying real and fake data. This process resembles a race, with the Generator and Discriminator continually chasing each other, ultimately reaching a balanced state where the generated data becomes increasingly realistic. Over time, many improved versions of GAN have been proposed to address issues such as training instability and mode collapse. Some of these enhancements include WGAN [16] (Wasserstein GAN), WGAN-GP [10] (WGAN with Gradient Penalty), CycleGAN [17], StarGAN [18], BigGAN [19], CoMoGAN [33], Inout-GAN [35] and more. These improved versions have achieved significant achievements in fields like image generation, style transfer, super-resolution reconstruction, and beyond.

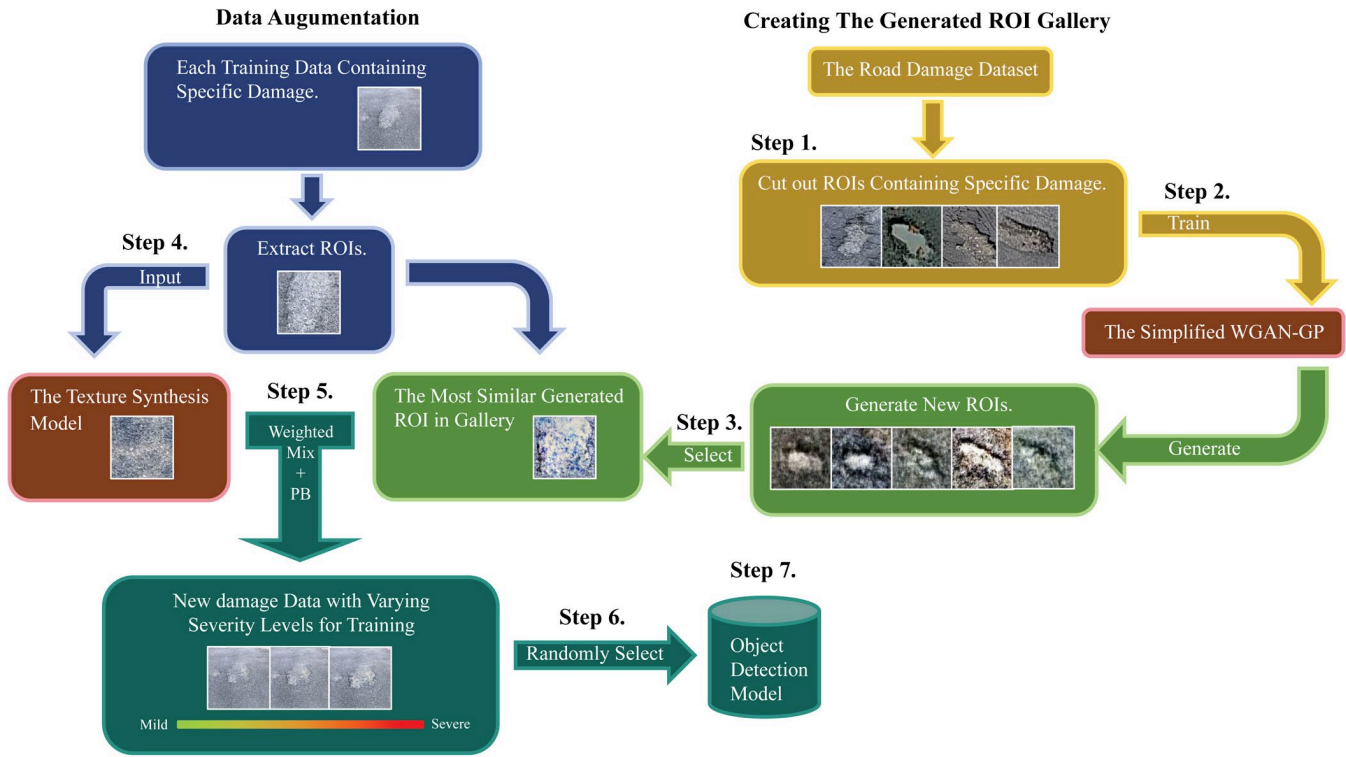

**Figure. 3** The overall process framework of the proposed method. "PB" stands for Poisson blending

Due to the potent generative capabilities of GANs numerous endeavors have harnessed them as a robust tool for data augmentation. For example, Bowels *et al.* [20] utilized synthetically generated samples with the appearance of real images to extract additional information from the dataset. They introduced GAN-derived synthetic data into the training dataset for two brain segmentation tasks, resulting in a 1-5 percentage point enhancement in Dice Similarity Coefficient (DSC). Han *et al.* [21] focused on exploring the utilization of GANs to generate synthetic multi-sequence brain magnetic resonance (MR) images. Their approach reached a level where even expert clinicians were unable to accurately distinguish between synthetic images and real samples in visual Turing tests. Similarly, Frid-Adar *et al.* employed GAN-based data augmentation techniques for liver lesion classification tasks, demonstrating that using GANs for data augmentation can achieve superior improvements compared to traditional augmentation methods. Zhou *et al.* [34] proposed IA-GAN, a method that adjusts the illumination level of training data to transform daytime data into nighttime data, thereby supplementing the data required for nighttime vehicle detection. This innovative approach successfully addresses the shortage of nighttime data in vehicle detection, significantly improving the model's overall performance in vehicle detection throughout the day. In the field of road damage detection, as mentioned above, Maeda *et al* [7] and Zhang *et al* [8] have effectively enhanced detection accuracy through GAN-based data augmentation methods in road damage detection. However, to achieve a more realistic quality in the synthesized data, their process still requires a significant investment of manual effort. Furthermore, their method primarily focuses on expanding the horizontal diversity of damage while neglecting vertical diversity. In certain cases, the introduction of synthetic data has even led to a decrease in accuracy in [7].

*C. Texture Synthesis*

Texture synthesis is a significant research area in computer graphics and computer vision, aiming to generate new texture images that visually resemble real textures. The fundamental principle of texture synthesis involves extracting samples from existing texture images and reassembling them to create new texture patterns. This process entails learning statistical features from the original texture, such as color, texture direction, scale, etc., and then utilizing these features to synthesize novel textures. This synthesis can be achieved through various methods, including statistical-based approaches [22], example-based methods [23], machine learning-based approaches [24], and deep learning-based approaches [11]. Texture synthesis finds extensive applications across various fields. In computer graphics, it can be used to generate realistic texture maps for rendering and simulation in virtual environments. In computer vision, texture synthesis can aid in image restoration by recovering missing or damaged texture regions within images. Furthermore, texture synthesis is employed in image style transfer, where texture features from one image are applied to another, achieving artistic effects. The versatility of texture makes it a valuable tool in these domains. In this paper, we employ texture synthesis to extract texture from the damaged regions of road surface. The obtained texture images are then combined with generated damage ROI through weighted mixing, enhancing the alignment between ROI and the background. Furthermore, during the mixing process, the parameter adjustments of weight allow us to control the severity degree of synthesized damage, which enables us to create varying degrees of severity for each damage in the dataset, thereby enhance the vertical diversity of synthesized data. The introduction of multi-gradient data distributions contributes to bolstering the model's robustness.

## III. METHODS

The core of our method lies in utilizing the texture synthesis model as an auxiliary method to extend the vertical dimension and improve the realism of the damage generated by the GAN. We combine the extracted road texture with the generated damage through weighted mixing. This enables us to achieve a better fit with the background, and by adjusting the weighting coefficients, we can control the severity of the synthetic damage, which benefits both realism and vertical diversity. We then embed the synthesized damage seamlessly into the original image via Poisson blending to ensure natural and smooth transitions at the edges. Additionally, we incorporate structural similarity to automatically select suitable synthetic damage during embedding, guaranteeing the quality of synthesized data while reducing significant manual effort. Finally, we employ a simple random selection strategy to mitigate distribution drift issues caused by an abundance of similar data. The overall framework of the proposed method is illustrated in **Fig. 3**. It consists of seven steps.

**Step 1.** Extract rectangular bounding boxes containing specified damage (pothole in our experiments) from the dataset.

**Step 2.** Train the simplified WGAN-GP model with the extracted ROIs as input. Then, utilize the trained generator to produce a substantial number of new ROIs, forming a generated ROI gallery.

**Step 3.** For each original data containing the specified damage, extract its existing ROI based on annotations. Subsequently, for each ROI, employ a script to iterate through the generated ROI gallery. During this process, compare the structural similarity between the original ROI and each generated ROI. This comparison aids in selecting the generated ROI with the highest similarity score.

**Step 4.** For the original ROI mentioned in (3), input it into the texture synthesis model to extract textures, obtaining road surface textures corresponding to their respective regions.

**Step 5.** Mix the generated ROIs selected in (3) with the textures extracted in (4) using weighted mixing. The weighting coefficients are adjustable, which results in controllable severity levels of road damage. Subsequently, employ Poisson blending to embed the synthesized damage back into the original ROIs' positions, resulting in new trainable data.

**Step 6.** To prevent distribution shift in the training data, we incorporate the synthesized data with varying severity levels obtained in (5) into the training set at random, following a predetermined ratio, for subsequent training.

**Step 7.** Train the object detection model on the expanded training set. We use the YOLOv5s for experimentation.

The details of our method are explained below.

*A. The Simplified WGAN-GP*

Due to the powerful generative capabilities of GANs, they have been extensively utilized for data augmentation to enhance model performance. To save training costs while ensuring the quality of generation, we employ a simplified WGAN-GP [10] as our generation model. As illustrated in **Fig. 3**, we utilize it to construct a generated ROI gallery.

WGAN-GP [10] introduces the Wasserstein distance, also known as Earth Mover's Distance, to replace the traditional GAN's Jensen-Shannon divergence (JS divergence). Traditional GANs utilize JS divergence to measure the difference between generated samples and real samples. However, JS divergence suffers from problems like gradient vanishing and gradient explosion during training, leading to instability. WGAN-GP employs Wasserstein distance, which has better mathematical properties and avoids the issues of JS divergence, making the training process more stable. The WGAN value function is constructed using the Kantorovich-Rubinstein duality [25] to obtain:

$$\min_G \max_{D \in \mathcal{D}} \mathbb{E}_{x \sim \mathbb{P}_r}[D(x)] - \mathbb{E}_{\tilde{x} \sim \mathbb{P}_g}[D(\tilde{x})] \quad (1)$$

Where $\mathcal{D}$ is the set of 1-Lipschitz functions and $\mathbb{P}_g$ is the distribution implicitly defined by $\tilde{x} = Generator(z). z \sim p(z)$ is the random noise.

Furthermore, to optimize the Wasserstein distance, WGAN-GP introduces a gradient penalty mechanism. By adding an additional loss term (similar to L2 regularization), the gradient of the discriminator is controlled within a reasonable range, further enhancing training stability. The specific approach involves calculating the linear interpolation between batches of real and generated samples during training, and then computing the gradients of the discriminator at these interpolation points. The difference between the norm of these gradients and 1 is used as the gradient penalty term, added to the discriminator's loss function. The specific formula for calculation is as follows:

$$\mathbb{E}_{\tilde{x} \sim \mathbb{P}_{\tilde{x}}}[(\|\nabla_{\tilde{x}} D(\tilde{x})\|_2 - 1)^2] \quad (2)$$

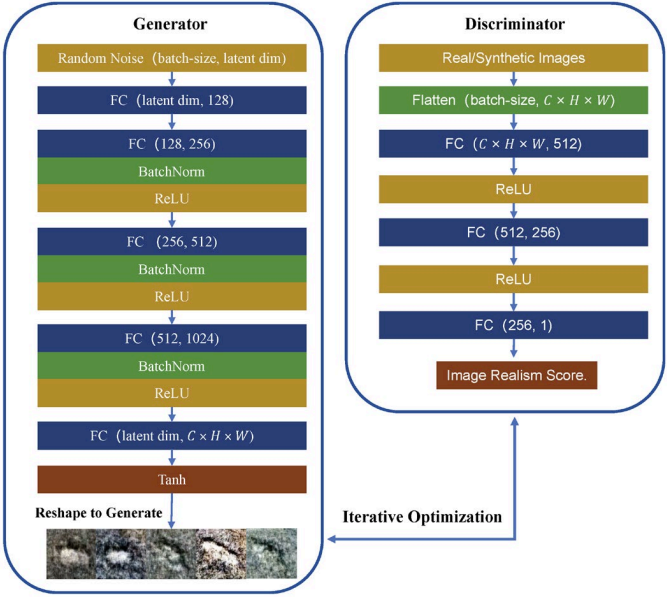

**Figure. 4** The network structure of the simplified WGAN-GP. "FC" stands for fully connected layer

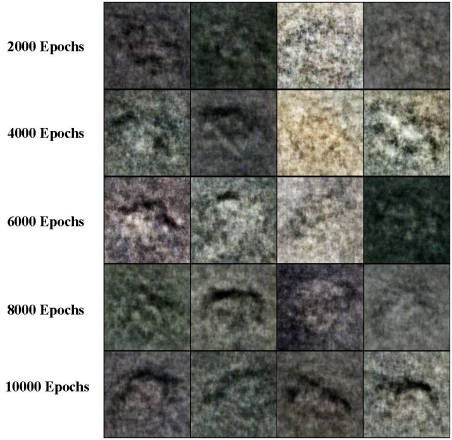

**Figure. 5** Potholes generated by the simplified WGAN-GP.

Therefore, the updated loss function is:

$$L = \mathbb{E}_{\tilde{x} \sim \mathbb{P}_g}[D(\tilde{x})] - \mathbb{E}_{\tilde{x} \sim \mathbb{P}_r}[D(x)]$$
$$+ \lambda \mathbb{E}_{\hat{x} \sim \mathbb{P}_{\hat{x}}}[(\|\nabla_{\hat{x}} D(\hat{x})\|_2 - 1)^2] \quad (3)$$

The original WGAN-GP utilized convolutional layers to extract image features, granting it strong generative capabilities even on datasets (e.g., LSUN Bedrooms [26]) with complex structures and intricate details. However, due to the simple structure and sparse nature of local details in pothole ROIs, we directly employed fully connected layers in place of conv layers

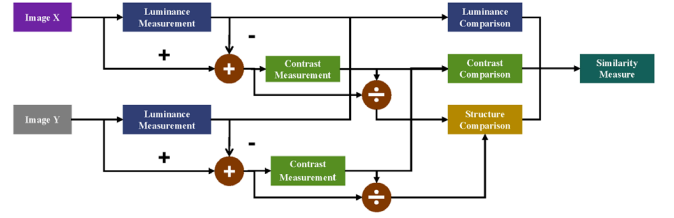

**Figure. 6** The diagram of structural similarity calculation.

The network architecture is illustrated in **Fig. 4**. The replacement operation reduces model parameters and computational load during training. Moreover, it efficiently generates synthetic pothole ROIs with pronounced damaged edge features after fewer iterations, as illustrated in **Fig. 5**.

We employ the trained generator to generate a substantial number of synthetic pothole ROI with diverse shapes and characteristics, constructing a generated ROI gallery to facilitate subsequent data augmentation, which ensures the horizonal diversity of generated damage.

*B. Selecting Appropriate Synthetic ROIs.*

While adversarial generation aids in creating a multitude of new ROIs, the crucial challenge lies in how to effectively select. Previous work, although successful in leveraging generated data to enhance the model, still encountered considerable manual intervention to some extent [7] [8]. They need to manually select the most suitable one from a large number of ROIs and embed it into the correct position. Our approach aims to completely liberate manual intervention from the data augmentation process and ensure that the embedded synthetic samples have a high level of adaptability. A fitting embedded sample should have relatively close visual angles, hues, and brightness levels to the original ROI. Therefore, it's necessary to design an appropriate matching mechanism.

For each original data containing the specified damage, we extract its existing ROI based on annotations. Subsequently, for each ROI, we employ a script to iterate through the generated ROI gallery. During this process, the machine compares the structural similarity between the original ROI and each generated ROI. This comparison aids in selecting the generated ROI with the highest similarity score.

Structural similarity is a commonly used metric to assess the similarity between two images. It takes into account factors such as structure, texture, and luminance. The schematic illustration of structural similarity is shown in **Fig. 6**, and its formula is as follows:

$$SSIM(x,y) = \frac{(2\mu_x \mu_y + C_1)(2\sigma_{xy} + C_2)}{(\mu_x^2 + \mu_y^2 + C_1)(\sigma_x^2 + \sigma_y^2 + C_2)} \quad (4)$$

Where $\mu_x$ and $\mu_y$ are the mean intensities, $\sigma_x$ and $\sigma_y$ are the standard deviations of images x and y, and $\sigma_{xy}$ represents the covariance. $C_1$ and $C_2$ are hyperparameters.

The machine-selected generated ROIs based on this criterion exhibit better compatibility with the original image background while also maintaining a certain level of distinctiveness from the original ROIs, as depicted in **Fig. 2**. This eliminates the need

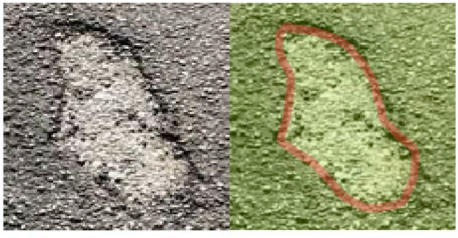

**Figure. 7** The pothole ROIs are rich in road texture features (green parts), while the pothole edge features (red parts) are discarded during the texture generation process for containing strong spatial information.

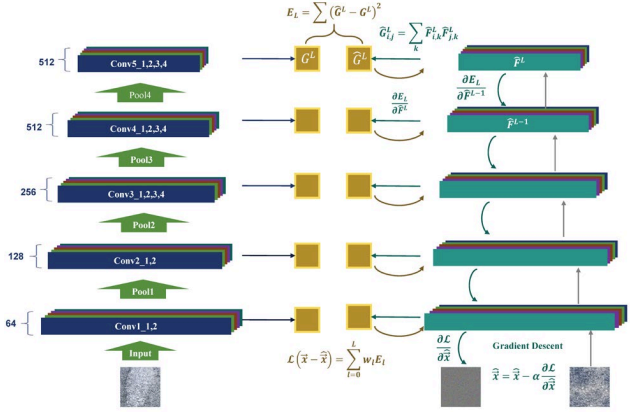

**Figure. 8** The network structure of the texture synthesis model

for manual selection entirely.

*C. Texture Synthesis*

Even though the machine-selected generated ROIs are comparable to manually selected ones, directly embedding them back into the original images might not yield satisfactory results, which neglect the severity levels of damage, and the realism still needs further improvement, as shown in the third column of **Fig. 2**. Therefore, we employed a texture synthesis model to extend the severity levels and improve the realism of the generated damage. In this section, we first elaborate on how we extract the texture of the road surface.

By observing the ROIs of pothole, we noticed that besides the outer edge of pothole, they also contain abundant road texture features, as shown in **Fig. 7**. With the help of a texture synthesis model, we can extract the road texture features and discard the pothole edge features. This allows us to synthesize road texture corresponding to the specified ROI regions. The texture synthesis model we employed has a structure as depicted in **Fig. 8**. It is a CNN-based texture synthesis model [11]. The model utilizes the VGG-16 [27] as the feature extraction network. It begins by uniformly extracting features of various sizes from the source image and calculates spatial statistics based on feature responses to obtain a stationary texture description. Finally, by performing gradient descent on a randomly initialized noise image, we generate a new image with the same stationary texture description.

Specifically, the input pothole ROI image is vectorized as $\vec{x}$, and then $\vec{x}$ is fed into convolutional neural network to compute the activation values at each layer, forming corresponding feature maps. If the $l_{th}$ layer has $N_l$ convolutional kernels, it generates $N_l$ feature maps. Let the size of these feature maps at this layer be denoted as $M_l$, and these feature maps are stored in a matrix $F_l \in R^{N_l \times M_l}$. Where $F^l_{j,k}$ represents the activation value of the $j_{th}$ convolutional kernel at position $k$ in the $l_{th}$ layer. As the texture description is stationary and independent of spatial information, this model discards spatial information from the feature maps and instead utilizes correlation coefficients among different feature maps to obtain summary statistic used for texture description. As a result, the pothole's edge features containing strong spatial information are discarded. This summary statistic is represented by the Cramer matrix $G^l$.

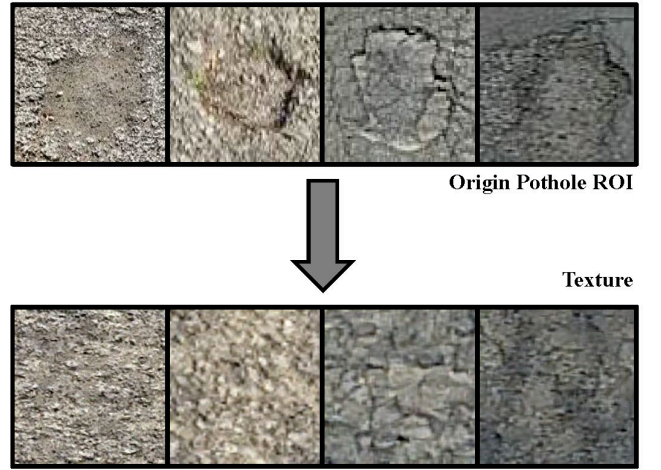

**Figure. 9** Demonstration of texture synthesis on pothole ROIs

$$G^l_{i,j} = \sum_k F^l_{i,k} F^l_{j,k} \qquad (5)$$

For a model structure with L layers, by computing the Cramer matrix for each layer, the complete texture description corresponding to the source image can be obtained, which is capable of uniquely characterizing a specific texture. The model uses random noise and employs gradient descent to generate a new image with the same texture. The underlying principle is to make the texture description of the generated image approach that of the source image. This process can be achieved through iterations, minimizing the average squared distance between the Cramer matrices of the source image and those of the generated image. Let $x$ and $\hat{x}$ represent the source image and the generated image, respectively. $G^l$ and $\hat{G}^l$ are their respective Cramer matrices at the $l_{th}$ layer. Therefore, the texture dissimilarity loss function at the $l_{th}$ layer can be expressed as follows:

$$E_l = \frac{1}{4N_l^2 M_l^2} \sum_{i,j} \left( G^l_{i,j} - \hat{G}^l_{i,j} \right)^2 \qquad (6)$$

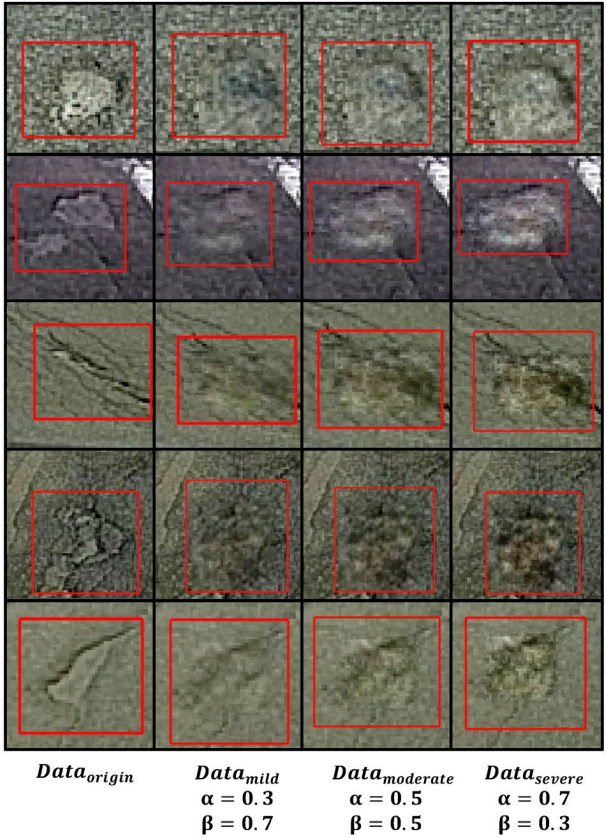

$Data_{origin}$ | $Data_{mild}$ $\alpha = 0.3$ $\beta = 0.7$ | $Data_{moderate}$ $\alpha = 0.5$ $\beta = 0.5$ | $Data_{severe}$ $\alpha = 0.7$ $\beta = 0.3$

**Figure. 10** $ROI_{mixed}$ with different parameter settings.

The overall loss function is as follows:

$$L(x, \hat{x}) = \sum_{l=1}^{L} w_l E_l \quad (7)$$

Where $w_l$ is the weight factor contributing to the total loss function for each layer. The partial derivative of $E_l$ is:

$$\frac{\partial E_l}{\partial \hat{F}_{ij}^l} = \begin{cases} \frac{1}{N_l^2 M_l^2} \left( (\hat{F}^L)^T (G^l - \hat{G}^l) \right)_{ji} & if\ \hat{F}_{ij}^l > 0 \\ 0 & if\ \hat{F}_{ij}^l < 0 \end{cases} \quad (8)$$

The model utilizes the L-BFGS [28] for iterative processing, which can complete the texture synthesis within a reasonable time using GPU and optimization tools for training. Given the relatively simple road texture features in the pothole ROI, we are able to extract its texture with just ten iterations, as shown in **Fig. 9**.

*D. Weighted Mix*

Now, for each original pothole ROI, we have a corresponding matched generated ROI image as well as its associated road texture. This section will detail how we utilize them to enhance realism and extend vertical diversity of the synthesized data.

We perform a weighted mix of these two elements to obtain $ROI_{mixed}$.

$$ROI_{mixed} = \alpha ROI_{generated} + \beta Texture \quad (9)$$

From **Fig. 2**, it is visually evident that the $ROI_{mixed}$ appear more naturally integrated into the original images compared to the non-mixed ones. Furthermore, by adjusting the weighted parameters $\alpha$ and $\beta$ we can manually control the severity of damage in the $ROI_{mixed}$. When $\alpha$ is large and $\beta$ is relatively small, the severity of damage in $ROI_{mixed}$ is higher, exhibiting prominent features. While $\beta$ is large and $\alpha$ is relatively small, the severity of damage in $ROI_{mixed}$ is lower, resulting in data that can be considered as challenging examples for the model. This is beneficial for enhancing the model's ability to identify damage with milder severity, thereby increasing recall. The $ROI_{mixed}$ under different parameter settings are illustrated in **Fig. 10**, it can be observed that the severity of the damage becomes more diverse.

Through weighted mixing, we obtained three different versions of $ROI_{mixed}$, labeled as $ROI_{mild}$, $ROI_{moderate}$ and $ROI_{severe}$. We label the data synthesized by embedding these three types of ROIs back into the original images through Poisson blending [12] as $Data_{mild}$, $Data_{moderate}$, and $Data_{severe}$, respectively. However, it is not reasonable to directly include all three versions of these in the training dataset. This is because these data have high overall similarity, with only the ROI region being different from the original images. Doing so is equivalent to adding four highly similar data to the training dataset, which is likely to lead to distribution drift problems. This has been confirmed by the experimental results presented later in the study.

To address this issue, for each training data, we randomly select one of its three augmented versions to add to the training set, with each version having an equal probability of being chosen.

## IV. Experiments

*A. Experimental Settings*

1) **Datasets**

RDD2019 (Road Damage Dataset 2019) [14] is a computer vision research dataset designed for studying road damage and maintenance. The objective of this dataset is to assist machine learning algorithms and computer vision models in identifying and classifying different types of road damage, facilitating better road maintenance and repair efforts. This dataset was collaboratively created by multiple research institutions and organizations, including public agencies in Japan, universities, and researchers in the relevant field. It consists of thousands of high-resolution images that capture various types of road damage, such as cracks, potholes, blur, etc. The images also encompass road conditions under various weather conditions to enhance the model's robustness. Most of the image in the dataset have been manually annotated, and the annotation information includes the type of damage, its location, size, and other relevant details. This annotation is crucial for training and evaluating machine learning models on the dataset.

We obtained the RDD2019 dataset with 10,186 annotated images. We initially divided these data into a training set and a validation set using an 8-2 split. The training set contains 8,148 images, and the validation set contains 2,038 images.

These data encompass seven categories, including 'D00,' 'D10,' 'D20,' 'D40,' 'D43,' 'D44,' and 'D50.' Among these, 'D40' represents potholes. Due to its relatively small dataset size with only 1,164 images in the training set and lower accuracy, we specifically applied data augmentation techniques to this category to evaluate the effectiveness of our approach.

2) **Experimental Configuration and Evaluation Indexes**

The experimental environment of this study was Ubuntu 16.04 + CUDA 11.7 + Pytorch 2.0.1 + Python 3.8, and the experimental device was an NVIDIA GeForce GTX 1080Ti. We employed YOLOv5s as our main object detection model. For the hyperparameter settings of YOLOv5s, the initial learning rate was set to 0.01, the learning rate strategy was cosine, the training cycle was 400, the initial input size of the model was 640×640, and the batch size was 8. For the simplified WGAN-GP, the initial learning rate was set to 0.0002, the training cycle was 10000. For the texture synthesis model, the training cycle was 10. For structure similarity, $C_1 = 0.01, C_2 = 0.03$. The mean average precision (mAP), precision (P), recall (R) and F1 score were chosen as evaluation metrics. A fixed intersection over union (IoU) value of 0.5 was chosen to calculate mAP in this study. Since we only performed data augmentation on the 'D40' class, which represents potholes, we will only compare the Evaluation Indexes for this class.

*B. Comparative Experiments*

We first conducted comparative experiments with the method [7], [32] and [38]. [7] uses PGGAN to generate new pothole ROIs and embeds them into suitable locations on pothole-free road surfaces using Poisson blending to create new training data. Since we cannot access their synthesized data and cannot confirm the specific codebase of their models, in this case, we directly compared the improvement in the F1-score when using the augmented model with the complete training set.[32] and [38] are currently considered practical general image data augmentation methods that take into account both cost and performance improvement effects. To ensure fairness in the comparison, we also employed the SSD-MobileNet and SSD-ResNet50 as utilized in [7] to evaluate the enhancement effects of these augmentation methods. as shown in TABLE I. The formula for calculating the improvement rate is as follows:

$$Rate_{improve} = \frac{F1_{augmented} - F1_{origin}}{F1_{origin}} \quad (10)$$

It's worth noting that the method in [7] requires a significant amount of additional manual labor, and it even has a negative effect on the relatively accurate model SSD ResNet50 [4]. The enhancement is also limited with only 2% improvement in F1-score when using the less accurate model SSD MobileNet [29]. The improvement effects of [32] and [38] on various detection models are relatively limited. In contrast, our method requires no human involvement and can be completed using simple scripts. Furthermore, it has achieved best enhancement results with different object detection models. For SSD-MobileNet, SSD-ResNet50, and YOLOv5s, our augmentation method has improved the F1-score by 6.1%, 7.9%, and 4.5%, respectively.

TABLE I
THE IMPROVEMENT COMPARISON OF DIFFERENT METHODS ON RDD2019.

| Method | $F1_{origin}$ | $F1_{augmented}$ | $Rate_{improvement}$ (%) |
|---|---|---|---|
| [7] with SSD MobileNet | 0.390 | 0.410 | 5.1 |
| [7] with SSD ResNet50 | 0.640 | 0.610 | -4.6 |
| [38] with SSD MobileNet | 0.423 | 0.428 | 1.1 |
| [38] with SSD ResNet50 | 0.614 | 0.616 | 0.3 |
| [38] with YOLOv5s | 0.627 | 0.623 | -0.6 |
| [32] with SSD MobileNet | 0.423 | 0.441 | 4.3 |
| [32] with SSD ResNet50 | 0.614 | 0.625 | 1.8 |
| [32] with YOLOv5s | 0.627 | 0.628 | 0.2 |
| **ours with SSD MobileNet** | 0.423 | 0.484 | **14.4** |
| **ours with SSD ResNet50** | 0.614 | 0.693 | **12.8** |
| **ours with YOLOv5s** | 0.627 | 0.672 | **7.2** |

Meanwhile, a comparison of the actual detection results after training the YOLOv5s on the augmented dataset and the original dataset is presented in **Fig. 11**. It is evident that after training on our augmented dataset, the model significantly reduces instances of missed detections, demonstrating higher detection quality. We have also visualized heatmap results for the model's detection, as shown in **Fig. 12**. It can be observed that the model trained on the augmented dataset is more sensitive to the pothole.

*C. Ablation Study*

To further analyze the effectiveness of our method, we conducted ablation experiments on the RDD2019 dataset. It can be visually observed from **Fig. 2** and **Fig. 10** that the data synthesized through our method possesses a more natural and realistic appearance and exhibits richer vertical diversity. So, how does the actual training performance when incorporating our augmented data into the training set? We tried several different data augmentation methods:

a. Adding only data that has not been blended with road textures, like the data in the third column of **Fig. 2**.
b. Adding only $Data_{mild}$.
c. Adding only $Data_{moderate}$.
d. Adding only $Data_{severe}$.
e. Adding all $Data_{mild}$, $Data_{moderate}$ and $Data_{severe}$.
f. Randomly selecting and adding $Data_{mild}$, $Data_{moderate}$ and $Data_{severe}$ in a 1:1:1 ratio.

The TABLE II shows that various enhancement methods have achieved improvements compared to the baseline. When we only add $Data_{mild}$, the model achieves the highest recall 0.664. This is because the addition of data B is equivalent to adding challenging samples to the training, which helps the model discover inconspicuous damages, thus improving recall. When we added all three types of data to the training set, the precision reached its highest point at 0.715. However, it's worth noting that the model's recall suffered a significant decline. **Fig. 13** presents the mAP results of the model across all categories for three scenarios: without data augmentation, using method-e, and method-f, from which we can observe that method-e led to distribution drift, affecting the model's performance in other categories and causing an overall decrease by 0.9% in mAP.

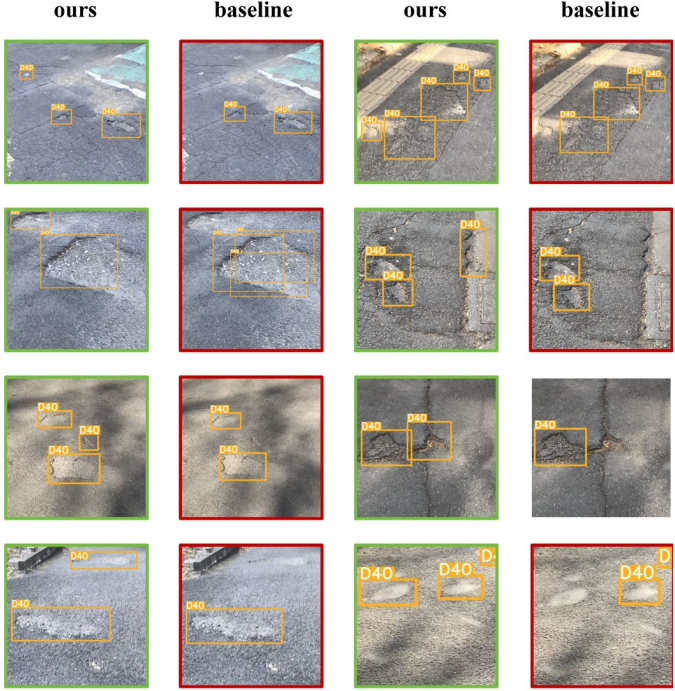

**Figure. 11** Quality comparison of detections: The green boxes illustrate the actual detection results of the YOLOv5s trained on our augmented dataset, while the red boxes display the actual detection results of the model trained on the original, non-augmented dataset.

TABLE II
ABLATION EXPERIMENTS ON RDD2019.

| Method | P | R | F1 | mAP |
|---|---|---|---|---|
| baseline | 0.622 | 0.633 | 0.627 | 0.637 |
| a | 0.645 | 0.627 | 0.635 | 0.657 |
| b | 0.64 | **0.664** | 0.652 | 0.663 |
| c | 0.672 | 0.639 | 0.654 | 0.655 |
| d | 0.666 | 0.632 | 0.649 | 0.657 |
| e | **0.715** | 0.598 | 0.651 | 0.647 |
| f | 0.695 | 0.65 | **0.672** | **0.678** |

Taking all factors into account, adding these three types of data to the training set after randomly selecting them in equal proportions is a better choice. In this way, the model achieved the highest F1-score 0.672 and mAP 0.678 in the pothole category, with improvements of 4.5% and 4.1% compared to the baseline, respectively. Furthermore, the model's performance in other categories was not significantly affected, even the overall mAP has improved by 0.4%.

## V. CONCLUSION

In this paper, we propose a novel data augmentation method for road damage detection. Our aim is to synthesize high-quality training data with diverse severity levels efficiently and reduce the need for manual effort and time. The overall process is not overly complex and can be applied to other specific tasks with similar characteristics to road damage detection. It's a versatile and effective method. The framework is highly extensible and offers room for improvement. For instance, the generative adversarial networks and texture synthesis models can be fine-tuned or replaced with more advanced models to further enhance data quality. Additionally, the strategy for randomly selecting augmented data can be refined for a more rational data selection process, which is also an area we plan to continue exploring and refining in the future.

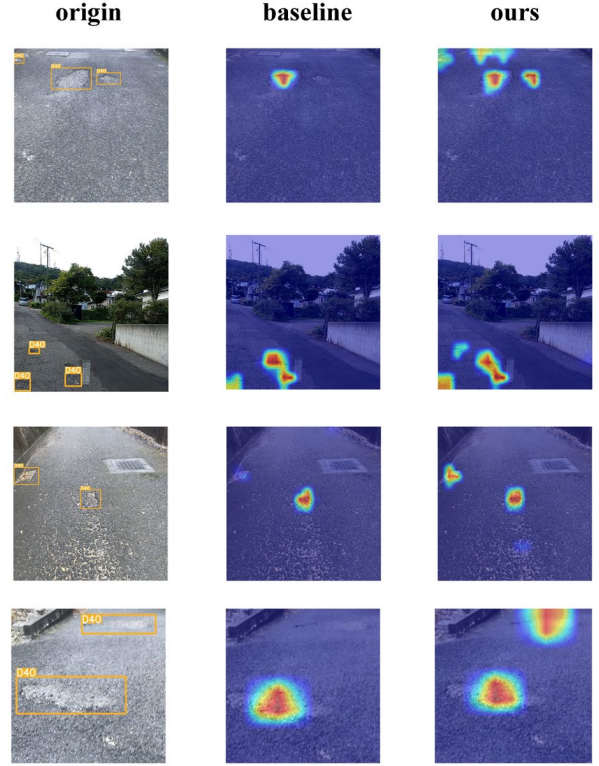

**Figure. 12** Visualization of heatmap results for some detection.

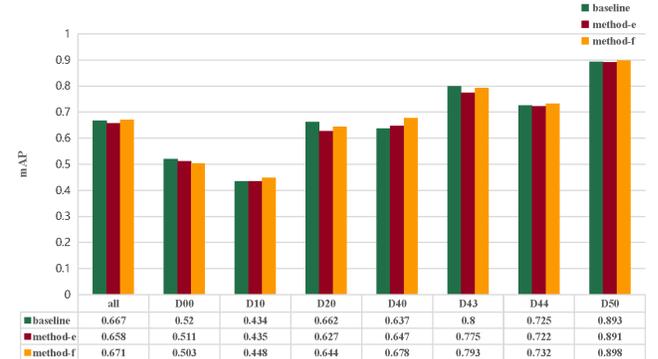

**Figure. 13** Comparison of mAP across all categories using methods-e and -f

| | all | D00 | D10 | D20 | D40 | D43 | D44 | D50 |
|---|---|---|---|---|---|---|---|---|
| baseline | 0.667 | 0.52 | 0.434 | 0.662 | 0.637 | 0.8 | 0.725 | 0.893 |
| method-e | 0.658 | 0.511 | 0.435 | 0.627 | 0.647 | 0.775 | 0.722 | 0.891 |
| method-f | 0.671 | 0.503 | 0.448 | 0.644 | 0.678 | 0.793 | 0.732 | 0.898 |

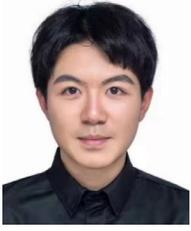

**First A. TENGYANG CHEN**, was born in Shantou, China, in 2000. He received the bachelor's degree in information & computing science from Sun Yat-Sen University. in 2022. He is currently pursuing the master degree in computer technology with Sun Yat-Sen University. His current research interests include intelligent transportation systems and image processing.

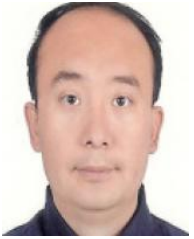

**Second B. JIANGTAO REN**, received the bachelor's degree in 1998 and the PhD degree in 2003 from Tsinghua University. He is currently an Associate Professor with the School of Computer Science and Engineering, Sun Yat-Sen University. His research interests include data mining and knowledge discovery, machine learning and intelligent transportation systems